\documentclass{article}

\usepackage{arxiv}
\usepackage[T1]{fontenc}    
\usepackage{hyperref}       
\usepackage{url}            
\usepackage{booktabs}       
\usepackage{amsfonts}       
\usepackage{microtype}      
\usepackage{graphicx}
\usepackage{bookmark}
\graphicspath{ {./images/} }

\title{Long-Term Planning with Deep Reinforcement Learning on Autonomous Drones }

\author{ \href{https://orcid.org/0000-0002-5911-0197}{\includegraphics[scale=0.06]{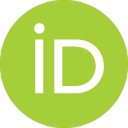}\hspace{1mm}Ugurkan Ates}\thanks{Evaluation run test video can be seen \href{https://www.youtube.com/watch?v=3wT8hmBlsRo}{at YouTube }} \\
	Department of Computer Science\\
	Gebze Technical University\\
	Istanbul, Turkey \\
	\texttt{ugurkanates97@gmail.com} \\
}

\hypersetup{
pdftitle={Long Term Planning with Deep Reinforcement Learning on Autonomous Drones},
pdfsubject={q-bio.NC, q-bio.QM},
pdfauthor={Ugurkan ~Ates},
pdfkeywords={Deep Reinforcement Learning,Path Planning,Long Term Planning,Autonomous Drones,Deep Learning,Machine Learning},
}

\begin{document}
\maketitle
\begin{abstract}
	In this paper, we study a long-term planning scenario that is based on drone racing competitions held in real life. 
	We conducted this experiment on a framework created for "Game of Drones: Drone Racing Competition" at NeurIPS 2019. 
	The racing environment was created using Microsoft's AirSim Drone Racing Lab. 
	A reinforcement learning agent, a simulated quadrotor in our case, has trained with the Policy Proximal Optimization(PPO) algorithm was able to successfully compete against another simulated quadrotor that was running a classical path planning algorithm. 
    Agent observations consist of data from IMU sensors, GPS coordinates of drone obtained through simulation and opponent drone GPS information.
    Using opponent drone GPS information during training helps dealing with complex state spaces, serving as expert guidance allows for efficient and stable training process. 
    All experiments performed in this paper can be found and reproduced with code at our GitHub repository
\end{abstract}


\section{Introduction}
\emph{Deep Learning} methods are replacing traditional software methods in solving real-world problems. 
Cheap and easily available computational power combined with labeled big datasets enabled deep learning algorithms to show their full potential. 
AlexNet paper(2012; Krizhevsky et al.\cite{NIPS2012_4824}) showed  feeding sufficient data into deep neural networks successfully learned to extract representations better than 
handcrafted features which let the start an era known as the rise of Deep Learning. 
Their great success in solving otherwise hard engineering problems such as  
\emph{object detection , voice recognition, chatbots, robotic manipulation and autonomous systems} 
shown they can be applied to various fields thanks to their generalisation capability.\cite{Simeone2018}

Path Planning(Motion Planning) is defined as
computing a continuous path from starting position S to destination position D while avoiding any known obstacles in the way.\cite{motion2020}
Whether it is in\textit{ 2D or 3D geometry}, any robotic system then will able to follow the computed path to reach it's destination.
Real World robotic systems tend to use more explainable and reproducible algorithms based on\textbf{ interval based search} ( \textit{A star or Dijkstra}) or \textbf{sampling-based }algorithms.
We wanted to show a reward based algorithm that depends on \textbf{Markov Decision Process}(\textbf{MDP}) by trying to maximize cumulative future rewards can also complete long term path planning tasks.
Advantage of using this option will allow autonomous robot(in our case simulated quadrotor) to create paths in \textit{non holonomic constraints} which is something current methods fails to achieve.\cite{Aranibar}\cite{SatinderSingh}

Deep reinforcement learning (RL) has been demonstrated to effectively learn a wide range of reward driven skills, \textit{including playing games} (Mnih et al., 2013)\cite{Mnih2013},\textit{ controlling robots} (2017; Schulman et al., 2015b)\cite{Schulman2017} and \textit{navigating complex environments} (Lei et al., 2017).\cite{Lei2018}

\begin{figure}
	\centering
	\fbox{\includegraphics[width=8cm]{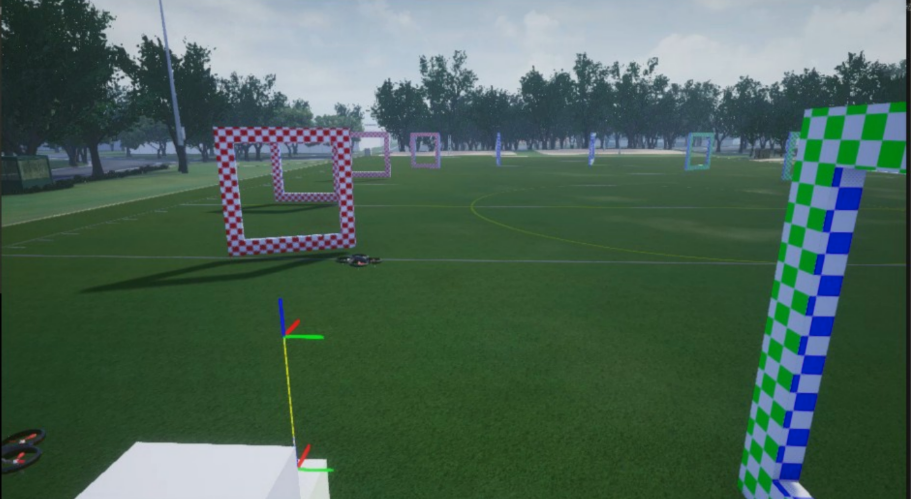}}
	\caption{An overview of simulated racing track.}
	\label{fig:fig1}
\end{figure}

We wanted to show an autonomous quadrotor acting as an agent in a fully known \textbf{MDP} can successfully learn to plan the optimized path on simulated drone racing track. 
Our agent was trained with \textbf{PPO} a very well-known once state of art continuous control algorithm that is still used due ease to implement and reproducible with few hyperparameters possible compared to other complex algorithms.

Our result shows after a fully trained agent could easily gain the ability to plan a long term planning scenario within a simulated racing track which consists passing ten of checkpoints. 
We also deployed our trained agent into similar conditions and it was able to generalize its path planning skills efficiently.

This shows promising results in the usage of drones in path planning scenarios as well as the deployment of these trained policies into real life.
Our sensor data is highly configurable and one can replace simulated sensors with actual sensor data. With proper calibration, trained policies can be deployed into real life systems.

\section{Preliminaries}

\label{sec:Machine Learning}
\paragraph{Machine Learning}
Machine Learning is in most basic form can be described as acquiring domain knowledge with a large number of examples of the desired objective.

These examples are called training set and fed into a learning algorithm of interest. Evaluated on test data they show the success rate of the training process.  
Learning algorithms try to maximize the success rate to reduce the disparity between predictions and actual outputs.\cite{Simeone2018}\cite{Mnih2013}

\label{sec:Reinforcement Learning}
\paragraph{Reinforcement Learning}
Reinforcement Learning is in the middle of supervised and unsupervised learning.
Unlike unsupervised learning  methods we have some form\textit{ guidance supervision system }but 
this doesn't affect the output of our desired output predictions instead, it acts like a 
feedback signal that tries to steer our policy to an optimal one, 
improving with each update to collect maximum cumulative reward.\cite{Henderson2019}

\label{sec:Deep Reinforcement Learning}
\paragraph{Deep Reinforcement Learning}
Deep Reinforcement Learning (DRL) has achieved great success on sequential task problems involving high dimensional sensory inputs such as RGB images or actual physical sensor data on real life robotic systems. Working on with highly dimensional input data there is some feature extraction layer that extracts using neural networks. These abstracted features then later used on to approximate Q value. \textbf{Deep Q Networks (DQN)} update policy regarding to Bellman expectation equation which includes an approximation of Q(state, action) with a neural network.
Another approach is the directly optimizing policy which results in \textbf{Policy Gradient} methods. These methods result in more stable training but they fail to generalize into any complex problem.

Both critic and value use \textit{separate neural networks} during training, Critic is updated with loss regarding bellman expectation function just like DQN, while critic updates with gradient ascent to maximize its policy to take actions that resulted in good rewards.

PPO is an advantage based actor-critic algorithm that tries to be conservative with policy updates. Using\textit{ KL divergence} and \textit{clipped surrogate function} it can update multiple steps of update in one trajectory without worrying about policy fluctuating wildly\cite{Schulman2017}. TRPO another similar algorithm does this by enforcing a \textit{hard trust-region constrain}t which solved with second-order methods.  PPO does this by with first-order optimization its clipped objective. \cite{Graves2020}\cite{Henderson2019}

\label{sec:Path Planning}
\paragraph{Path Planning}
The motion planning problem is often related to the \textit{subtask graph execution} problem. 
Problem is often mapped to a graph and thus becoming a graph search problem. 
The key concept behind the problem is to solve finding an optimal path toward the target while avoiding obstacles in the way.\cite{motion2020}\cite{Cai2020}

\section{Related Work}
\label{sec:Hierarchical Reinforcement Learning}
\paragraph{Hierarchical Reinforcement Learning}
In recent years Hierarchical Reinforcement Learning (HRL) has started to show promising results,
solving \textit{long horizon problems} on environments with sparse or delayed rewards.
We could classify these algorithms into two type.
First algorithm type focuses on high level policies ability to choose low level actions which they are called \textbf{selector methods}
Second type of algorithms rather focuses on determining high level policies from low-level policies which are called \textbf{subgoal based methods.}

Both algorithm type can show enable transfer within different environments and can solve diverse tasks. Selector algorithms mostly

\textit{Selector methods} enable convenient skill transferring and can solve diverse tasks. They often require
training of high-level and low-level policies within different environments, where the low-level skills
are pre-trained either by proxy reward , by maximizing diversity , or in designed simple
tasks.\cite{Li2019}\cite{Yu2018}

\textit{Subgoal-based methods} are designed to solve sparse reward problems. A distance measure is required
in order for low-level policies to receive internal rewards according to its current state and the subgoal.
Many algorithms simply use Euclidean distance [8, 9] or cosine-distance  as measurements.\cite{Sohn2019}

\label{sec:Multi Task Learning}
\paragraph{Multi Task Reinforcement Learning}
Multi-Task Learning (MTL) tries a more general approach when dealing with a problem with multiple sub-tasks. Instead, it uses leveraging the power of Machine Learning to generalize upon each task given properly chosen hyperparameters and given enough training time it learns other subtasks. This method also increases the learning capacity of each subtask at once and improves the ability to learn several tasks. This is mostly achieved due to similarities of subtasks that could be learned from features extracted in hidden layers could be more efficiently exploited to learn other tasks and by a general representation of each subtask to learn.\cite{TungLongVuong}

Multi-Task RL(MTRL) methods could pass these efficiency improvements made in the Deep Learning domain to the RL domain by allowing multiple tasks to be learned by a single agent. The same domain example to this could be reaching race-walking and running, a similar domain example could be a single pendulum or double pendulum balancing task. \cite{Jin2017}

Authors  conducted an experiment to compare MTRL algorithms.\textbf{ Multi Deep Q Networks} with separate replay memories for each task learned ability to generalize on multiple tasks while having time complexity then Vanilla DQN. They also constructed to actor-critic domain with \textbf{Multi Deep Deterministic Policy Gradient} extended their ability to learn continuous action domains in the same vein, which they showed with pendulum task.

\label{sec:Path Planning with Reinforcement Learning}
\paragraph{Path Planning with Reinforcement Learning}
There are many works of motion planning with DRL  \cite{Aranibar,Lei2018,article,Cai2020},. Bae et al.  compared Depth-first methods with RL their results on experiments
and shown a RL algorithm shown paths generated by RL methods more easily executed by robots then Depth Methods, which generate a polygonal line paths
Paths generated by RL  almost in all situations achieve a low orientation error when executed on the real robot. \cite{article}

They also showed combined with local \textit{LIDAR} information a RL based \textit{path planning }was able to generalize into wider applications and providing
better generalization performance without getting trained for each unknown dynamic environment.

The problem of multi robot path planning is motivated by many practical tasks because of its efficiency for performing given missions. However, since each robot operates individually
or cooperatively depending on the situation, the search area of robot is increased. \cite{Cai2020}

Reinforcement learning in the robot's path planning algorithm is mainly focused on moving in a fixed space where
each part is interactive.
In most cases, existing path planning algorithms highly depend on the environment. \cite{Lei2018}

\label{sec:Drones with Reinforcement Learning}
\paragraph{Drones with Reinforcement Learning}
The works on Drones have long existed since the beginning of RL.\cite{drones3030072,Madaan2020,SatinderSingh} . Some the early work done was involving
\textit{dynamic programming} algorithms to some complete basic tasks with RL algorithm to show the capability of RL with drones but due lack of powerful function approximators such as neural networks they couldn't be used in complex real-life situations.

With Deep Learning revolution we have seen drones starting to use\textbf{ Deep Learning} based 
object detection, image recognition, etc. techniques to better navigate their paths and thus complete highly precision required tasks such as delivering cargo\cite{drones3030072} and creating\textit{ HD map}s of dangerous parts of the world.
This, of course, comes with a caveat, since DL methods are more compute hungry battery lives become a more apparent problem.

\section{Long Term Planning with  Deep RL on Autonomous Drones}

\subsection{How does it works}
We had access to modeled accurate representation of the actual drone racing track.\cite{Madaan2020} The track is highly customizable thanks to AirSim APIs,(see Section \ref{airsim}) this helped to easily modify parts of it in towards to achieve \textit{domain randomization}. We spawned our training agent in different locations during the training process for learning to recover it's the path when it went off during the evaluation process.

Our long term task is the complete racing track which consists of ten(10) mini-tasks to finishing faster than competitor drone. Our agent should try to avoid corners of gate checkpoints, hitting too many corners would distract our planning and lead us into an unrecoverable state.

Agent gets rewards signals for each step taken in the environment. \textbf{Reward engineering} is the process of creating appropriate rewards for learning agents to collect during training.\cite{Dewey2014ReinforcementLA,Eysenbach2018} Our technical details regarding details of explained in here.(See Section \ref{tech}). Completing all subtasks and finishing race defined as one episode of MDP process.PPO algorithm we used here can update its policy multiple times with one trajectory. We used competitor drone equipped the same physical appearance and aerodynamics to collect guidance data via API calls. A simple path planning algorithm has been used for competitor drone. Using an \textit{expert-guided approach} allows our agent to not stuck in local optima and away from the target goal. Especially with high dimensional state problems its very hard to train agents from scratch. Exploration in the environment mostly supplied with\textit{ expert guidance and hand-designed rewards. }

 Using a function approximator like neural networks we achieved near-real human operator performance with enough training time. 
We have explained our observations and actions in detail.(see Section \ref{obser} )

After 2.5 million steps taken in simulations, our agent started partially complete ten gate checkpoints while hitting some of the obstacles and stuck in local optima. 
Using the same network we have trained 4 million steps more training with the same seed. While continuing trained networks isn't much common in Deep RL domain we had to due limited computational power we experimented with. After starting of each continued training process agent had to adapt itself to new trajectories thus felt a performance drops were expected.

\label{airsim}
\subsubsection{AirSim Drone Racing Lab Framework}

\begin{figure}
	\centering
	\fbox{\includegraphics[width=12cm]{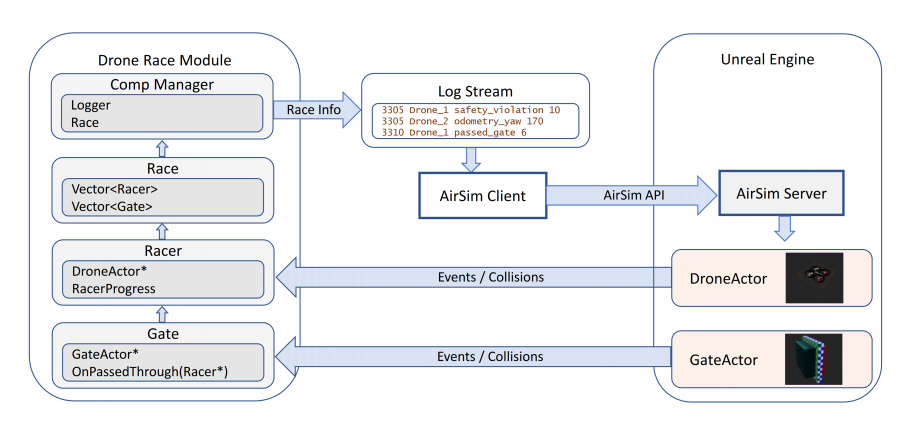}}
	\caption{General overview of communication between agent and simulation .}
	\label{fig:fig2}
\end{figure}
The goal of our AirSim Drone Racing Lab simulation framework is to help bridge the gap between simulation and reality by utilizing 
high fidelity graphics and physics simulation.\cite{Madaan2020} 
It uses\textit{ Epic's Unreal Engine 4 }for graphic rendering pipeline while physics engine is deployed into developed by themselves. The main goal of it enabling to train different
machine learning algorithms with the ability to create unique scenarios and conditions with given high level APIs to developers.
It features the ability to use their flight controller as well as replacing the default controller with something more sophisticated. It also features
ability to specify inertial sensors which would enable more realistic training sessions.

\textit{Drone racing} involves multiple competitor agents and due nature can
also be modeled as multi-agent problem. We have used an already available basic path planning algorithm implemented inside the simulation environment which helped us with training and evaluation processes. It is also fairly easy to import 3D models such as drone gates, obstacles to randomize domain. We can achieve most of the racing info with API calls which let us access monitor score and progress as well as time penalties. 

It is also fairly simple to get camera data located on the top and behind each drone available in the simulation. Ground truth data can be extracted from simulation which would ease the process of data labeling in an \textbf{Imitation Learning }approach.

\subsubsection{Proximal Policy Optimization} 
Proximal Policy Optimization is model free on-policy actor critic algorithm introduced by Open AI (Schulman et al. 2017)\cite{Schulman2017}
that tries to uses \textit{clipped surrogate function} as an optimization goal.

With it's easy to implement nature and requirement to adjust fewer parameters with 
comparable results to other on-policy algorithms,PPO has become one of the most frequently used continuous control problems.

\begin{equation}
\max _ \theta \mathop{\mathbb{E_T}} [\frac{\pi_\theta(a_t|s_t)}{{\pi_\theta}_{old}(a_t|s_t)}] A_t(s_t,a_t)
\end{equation}

Here \begin{math}A_t\end{math} stands for  generalized advantage function. (Schulman et al. 2015b)\cite{Schulman2017}

\begin{math}\varepsilon\end{math}  hyperparameter is used for \textit{ratio of clipping }between 1-\begin{math}\varepsilon\end{math} , 1+\begin{math}\varepsilon\end{math} depending on whether advantage is positive or negative.

It is widely believed that the key innovation of PPO responsible for its improved performance over the baseline of TRPO is the ratio clipping mechanism over KL divergence.

\label{obser}
\subsubsection{Observations and Actions} 
The RL-agent needs to get all relevant information about the current state of the environment to be able to fulfill the task successfully. The following listing provides the raw data
that has been identified as relevant.
In this work, we have used quadrotors \textbf{IMU}(\textbf{Inertial Measurement Uni}t) sensors which contains information of linear velocity,angular velocity,
gyroscope data for which contains \textit{pitch(x-axis), roll (y-axis), and yaw (z-axis). }
The gyroscope has no initial frame of reference (like gravity), you can combine its data with data from an accelerometer to measure the angular position.
We also used GPS data we were able to collect from APIs of AirSim. 
These provided a rich understanding of the environment thanks to realistic continuous state inputs. 

We also use our competitor drone equipped with a classical path planning algorithm also provide us its location which acts a guidance mechanism during training.

We have opted to use already implemented low-level drone PID controller - 
Since our focus was on path planning rather than improving on PID controller, though this also reminds a point for improvement in the future.

From our tests, we have seen need for clipping due to inefficient learning so we have used clipped versions of each sensor input data and outputs from 
our neural nets.
Our neural network outputs speed difference in terms of x,y,z coordinates each step given to controller API  which then allows our drone to balance and improve on.

\subsubsection{Reward Engineering} 
Reward engineering is a highly active research topic in the RL domain and researchers have come up with different ideas related to it. 
Inverse RL  tries to approximate reward function from expert demonstrations shown some success in small toy tasks but has yet to generalize for complex tasks. Handwritten reward engineering techniques remains a common practice when designing a RL system though problems such as low-sparse dilemma remain to be solved. In our experiment, we have handcrafted rewards and which was the most time-consuming part of this project. 

Our agent had to pass the checkpoint at gates without hitting collapses in \textit{3D geometric space}. We had to create a\textit{ feedback signal mechanism} so it wouldn't receive negative rewards for not getting the next gate after passing a gate. We created a timer function that would keep track of each gate time and decide rewards for that gate given expert demonstrations by our competitor drone.
 
Each training episode spawned at a different but constricted random location which had\textbf{ 2.0 -3.5 meters }into actual gate checkpoints, this allowed for more \textit{domain randomization}.

We feed agents with positive reward signals based on proportional to their distance its current target gate. Distance between each was approximately is 10-15 meters, and we had given extra positive rewards when the agent was in close range of 3 meters to target as a positive incentive. When an agent was 0.5 meters close to its current target gate we started calling our distance calculation function to properly calculate if our agent passes inside the gate.  

Our calculation was based on \textit{L2 distance} of the middle point of each gate proportional to drone orientation. We give a huge positive reward at the passing of each gate and start timer alongside with it. Until timer for each game runs out our agent has to leave its current area and to move towards the next target gate. We have given penalty rewards to our agent that stuck with the old target gate. Choosing the right values for this situation required a lot of trial and errors to find the proper fit.
\begin{figure}
	\centering
	\fbox{\includegraphics[width=9cm]{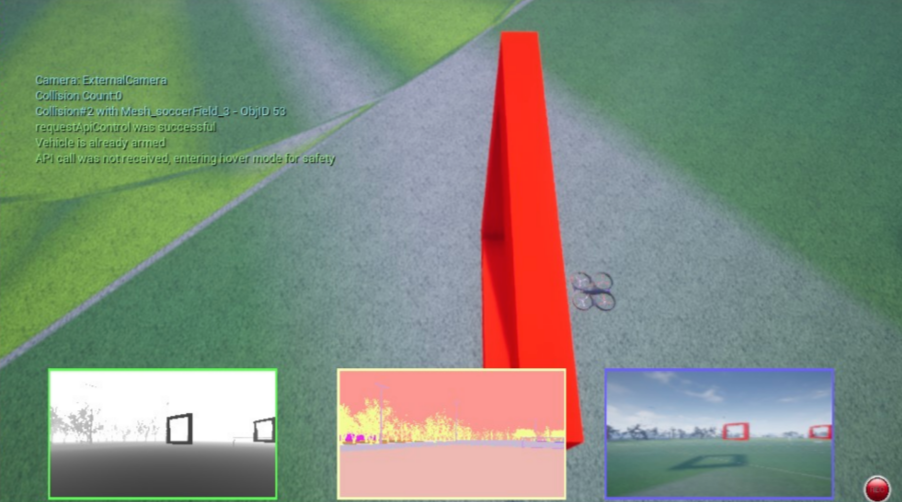}}
	\caption{A timer starts to incentive drone to reach next gate checkpoint.}
	\label{fig:fig3}
\end{figure}
We terminated episodes if the agent was going too far from its target destination. Any attempt on going too far from gate checkpoint that wasn't detected by us was crashing simulation and we could detect crashes in simulation thus allowed reset environment for next training run.We also kept track of time and number of hits collisions to gate corners for negative penalty signals.

\section{Experiment}

\label{tech}
\subsubsection{Implementation Details} 
Our neural network has \textit{3} layered \textit{MLP} outputting speed coordinates (x,y,z) respectively that should be inputted controller of quadrotor to increase or decrease the speed. We thought about using a\textit{ recurrent neural network} for allowing better memorisation but at first tests it didnt seem to improve much so we dropped for a more simplistic network.
Our neural network had hidden size of \textit{256} along with starting learning rate \textit{1e-4.}
Our trajectories were \textit{2048} steps and mini-batch consisted of \textit{256} transitions each time.

We clipped neural network output to between\textit{ -1 and 1.} In our experiment, this shows stability and convergence unlike first 
the approach we tried resulted  have failed results. 

We also normalized every state value and rewards for an efficient training process. 
We run the simulation at arbitrary\textbf{ 640x480 resolution at 60 frames per second} with all graphical effects in low, shadows disabled. Our training conducted in cloud GPU providers and we downloaded model weights regularly to see the running model in the local simulation. Unfortunately, \textit{Unreal Engine API} didn't allow us to record video on server machines. 

We have created a mechanism to stop training and continue as efficiently possible with minimal effect on pre-train drops
Communication with our model and simulation was made possible via \textit{TCP} communication built inside the simulation , we have also extended some of its abilities for remote monitoring of data packages.

\subsubsection{Evaluation and Results} 
Our goal was to show teach a RL trained agent to complete ten gate checkpoints track given time limits,
 minimizing hitting obstacles and being able to recover and continue the optimal path from given limits.In order to achieve this we have hand crafted a unique solution customized for drone racing environment which can be extended to similar domains. In our early tests we have seen bad effects of no clipping which led stuck in various local optima positions. In order to eliminate this issue we have normalized our state and action values , this way any in case of getting away from optimal path would result in big negative penalty but it was still recoverable due advantage policy update nature of algorithm.    
 
 \begin{figure}
	\centering
	\fbox{\includegraphics[width=9cm]{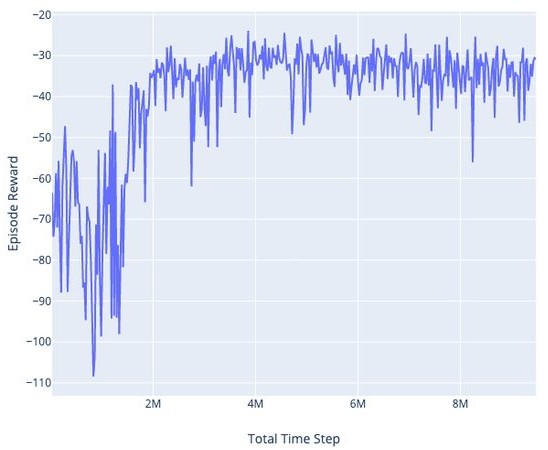}}
	\caption{Episodic Reward - Time Steps during training.}
	\label{fig:fig4}
\end{figure}
 We have recorded a video of the drone during the evaluation run and published our results as reproducible with network weights to test and improve upon.

\section{Conclusion and Future Work}
In this experiment, we have shown a fully trained reinforcement learning agent with using  AirSim Drone Racing 
simulation can achieve successful long term planning using continuous state values and continuous actions. 

\textbf{\textit{Deep Reinforcement Learning}} algorithms coupled with strong function approximators can achieve great results at path planning tasks as well as generalization of said tasks. Spawning agents from different starting positions achieve tries to achieve domain randomization for agents which then allows recovery from hard states. This has shown with deploying agents on different spawn points and seeing their ability to recover and continue at an optimal path. This work can be extended to real-life drone racing situations due to the similarity between 

We believe a properly calculated IMU data of actual drone can allow us to deploy trained policy into real life without having a hard time adapting.  Deploying trained simulation policies can allow training for a lot of interesting planning scenarios due to the ease of training agents in the simulation. One example could feature planning and delivery tasks for a  drone that carries packages in real life.

\section{References and Source Code}
\label{sec:others}

The documentation and source code for \verb+project+ may be found at
\begin{center}
	\url{https://github.com/ugurkanates/NeurIRS2019DroneChallengeRL}
\end{center}
\bibliographystyle{plain}\bibliography{references}
\end{document}